\documentclass[11pt,a4paper]{article}

\usepackage[T1]{fontenc}
\usepackage[utf8]{inputenc}
\usepackage[english]{babel}
\usepackage[expansion=false]{microtype}
\usepackage{amsmath,amssymb}
\usepackage{booktabs}
\usepackage{graphicx}
\usepackage{url}
\usepackage[round]{natbib}
\usepackage{geometry}
\usepackage[colorlinks=true,citecolor=black,linkcolor=black,urlcolor=blue,
  pdftitle={MISO-BR: An Open Corpus for Misogyny Detection in Brazilian Portuguese},
  pdfauthor={F\'abio A.S. Passos, P\'ericles B.C. Miranda, Ana Paula Cavalcanti},
  pdfkeywords={misogyny detection; Brazilian Portuguese; annotated corpus; hate speech; BERTimbau}]{hyperref}

\newcommand{\corpus}{MISO-BR}

\title{\textbf{\corpus{}: An Open Corpus for Misogyny Detection in Brazilian Portuguese}}

\author{
Fábio A.S. Passos \\
\small CESAR School
\and
Péricles B.C. Miranda \\
\small Departamento de Computação\\ UFRPE
\and
Ana Paula Cavalcanti\thanks{Corresponding author: \texttt{ap.cavalcanti@ufrpe.br}} \\
\small Departamento de Computação\\ UFRPE
}

\date{}

\begin{document}

\maketitle

\begin{abstract}
\noindent
Misogyny is among the most prevalent forms of online hate speech directed at Brazilian users, yet open and auditable resources dedicated to it in Brazilian Portuguese remain scarce. We introduce \corpus{}, an open, documented and versioned corpus of 896 balanced social media texts annotated for binary misogyny, released with its construction protocol and reproducible baselines. The corpus combines lexicon- and target-guided scraping from X with items imported from OLID-BR, a large language model triage stage whose prompt and decoding parameters are released, and manual annotation by three annotators following a written guide. Inter-annotator reliability reaches a Fleiss' $\kappa$ of $0.788$ over 816 items. Under a single fixed protocol we compare four classical classifiers (Multinomial and Bernoulli Naive Bayes, a linear SVM and a random forest) with fine-tuned BERTimbau Base and Large. The classical models cluster between $0.651$ and $0.676$ macro-$F_1$ and are precision-oriented but recall-poor. The transformers reach $0.833$ macro-$F_1$ and accuracy (Wilson 95\% CI for accuracy $[0.77, 0.88]$), a gap of roughly 16 points, while Base and Large are indistinguishable on the test set despite a $2.6\times$ difference in training cost. We argue that the value of the resource lies less in its headline scores than in the auditability of its pipeline, and we state three limits that users must reckon with: the artificial class balance, whose effect on precision under realistic prevalence we quantify; the association between provenance and label, which the released \texttt{source} field makes measurable; and the selection bias induced by lexicon-guided sampling and automatic triage. The corpus, annotation guide, models and experimental scripts are publicly available.

\vspace{0.6em}

\vspace{0.6em}
\noindent\textbf{Content warning:} this paper quotes misogynistic and otherwise offensive language for illustrative purposes.
\end{abstract}

\section{Introduction}

Two decades of expansion of social platforms have reorganised public discourse, and with it the distribution of who can speak and at what cost. Alongside the widening of participation, the same infrastructure has industrialised the circulation of hostility toward groups that were already marginalised offline, a phenomenon that has become one of the central preoccupations of both civil society and computational linguistics \citep{fortuna2018survey,poletto2021resources}. Gendered hostility occupies a distinctive place in this landscape. It is not simply hate speech that happens to have a woman as its object; it is a discursive mode with its own repertoire, its own justificatory grammar, and its own tendency to attack the speaker's legitimacy rather than her argument.

The Brazilian case is well documented and severe. The MonitorA reports show that digital environments reproduce and intensify patterns of discrimination present offline, and that women in positions of public visibility, journalists, politicians, researchers and athletes, face attacks that route around their ideas and target their gender \citep{monitora2024}. SaferNet Brasil received 23{,}947 reports of online hate speech in the first half of 2022, an increase of roughly $67.5\%$ over the same period of the previous year, and misogyny was the most reported category in absolute terms, with 7{,}096 reports \citep{safernet2022}. Volume, however, understates the problem. A substantial share of misogynistic speech travels in a jocular, sarcastic or ironic register, in which the offence is carried by presupposition rather than by any single lexical item, and it is precisely this share that automatic systems handle worst.

Progress on automatic detection is bounded by the resources available, and here Brazilian Portuguese occupies an awkward position. It is not a low-resource language in the usual sense, and it is not well served either. A sequence of corpora has established a foundation for offensive and hateful language in the language, from OffComBR \citep{pelle2017offensive} through the hierarchically labelled dataset of \citet{fortuna2019hierarchically}, ToLD-Br \citep{leite2020toxic}, HateBR \citep{vargas2022hatebr}, OLID-BR \citep{trajano2023olidbr} and TuPy-E \citep{oliveira2023tupye}. Almost all of them are generalist. Where gender appears, it appears as one category among many, usually labelled \emph{sexism}, defined within a schema whose primary axis is offensiveness rather than gender, and annotated by people whose instructions were written for the general case. ToLD-Br does carry an explicit misogyny label, but only 29 instances reached consensus among its three annotators, which is a count that supports an observation rather than an experiment. The first Brazilian corpus built specifically around hate speech against women, MINA-BR \citep{plath2022minabr}, was described in 2022, but its data were published in the research data repository of Unicamp only in March 2025 \citep{plath2025minabrdata}, with publicly downloadable files and no stated licence, and independent work on it has only recently begun to appear \citep{vicente2025detecting}. A more recent effort pairs a misogyny lexicon, LexMis-BR, with lexicon-guided collection and human annotation \citep{rangel2026misoginia}.

The consequence is a field in which specialised resources are only beginning to become inspectable. Until March 2025 the most direct antecedent to new work on Brazilian misogyny detection had not been publicly deposited, and the specialised resources that now exist differ in what they label, hate and offensiveness in the case of MINA-BR rather than misogyny as distinct from non-gendered toxicity, and in how much of their construction they expose. This is not a criticism of those efforts, whose contributions stand, but it is a structural problem: without open specialised resources whose pipelines can be audited there is no shared object against which annotation schemes can be argued about, no benchmark against which modelling claims can be checked, and no way to distinguish a genuine advance from a favourable test split.

This paper contributes \corpus{}, an open corpus for binary misogyny detection in Brazilian Portuguese, released with its annotation guide, its data dictionary, its licence, its versioning and its baselines. The corpus comprises 896 instances and rests its claim to relevance on three properties. First, it is specialised: every annotation decision was made under a definition of misogyny, not under a definition of offensiveness with a gender attribute attached. Second, it is auditable end to end, since the scraping strategy, the lexicons, the triage prompt and its decoding parameters, the annotator profiles, the disagreement cases, the resolution rule and the provenance of every instance are all documented and, where applicable, released. Third, its numbers are reported together with their limits. We report confidence intervals alongside point estimates, we do not rank models whose differences amount to four test items, we do not draw the cross-corpus comparison that the reported literature invites, and we quantify the degradation of precision that follows from evaluating a balanced corpus and deploying against a skewed stream.

Alongside the corpus we contribute a benchmark evaluation designed to establish what it supports. Six models are compared under one fixed protocol and one fixed partition, spanning the two paradigms against which the Brazilian literature has reported: four classical classifiers built on bag-of-words and TF-IDF representations, and two fine-tuned BERTimbau variants. The comparison is informative in both directions. It establishes the baselines a successor system must beat, and it locates where the difficulty of the task actually sits, since the classical models plateau at a recall near $0.60$, while the transformers recover a substantially larger share of the positives, for a difference of roughly $16$ macro-$F_1$ points. The evaluation also settles a practical question of scale, in that the additional capacity of BERTimbau Large yields no measurable gain over Base for $2.6\times$ the training cost.

The remainder of the paper is organised as follows. Section \ref{sec:related} situates \corpus{} in the international and Brazilian literature. Section \ref{sec:construction} describes the construction pipeline and characterises the resulting corpus. Section \ref{sec:setup} presents the experimental protocol, and Section \ref{sec:results} reports the results together with the qualifications they carry. Section \ref{sec:limitations} states what the corpus cannot support, and Section \ref{sec:conclusion} concludes.

\section{Related Work}
\label{sec:related}

The international literature separated misogyny from general abuse relatively early. The Automatic Misogyny Identification tasks at IberEval and EVALITA established the first shared evaluations dedicated to the phenomenon, with schemas that distinguished categories such as stereotype, dominance, derailing, sexual harassment and discredit, and that separated the misogyny judgement from the question of whether the target was an individual or women as a group \citep{fersini2018iberval,fersini2018ami}. The later MAMI task extended the problem to multimodal content, acknowledging that a large share of contemporary misogynistic circulation is memetic rather than textual \citep{fersini2022mami}.

Two subsequent contributions are directly relevant to the methodological posture we adopt. \citet{guest2021expert} built an expert-annotated English dataset with a hierarchical taxonomy of online misogyny and, crucially, reported that the phenomenon is rare enough in unfiltered data that meaningful corpora require deliberate sampling, which in turn distorts the resulting prevalence in ways that must be declared. \citet{zeinert2021annotating} made the annotation process itself the object of study for Danish, showing that the reliability of misogyny annotation depends heavily on how the guideline handles context, ambiguity and reclaimed slurs, and that a single aggregate $\kappa$ conceals systematic differences between annotator subgroups. Both findings inform how we present agreement in Section \ref{sec:construction}.

The broader abusive language literature has converged on a set of cautions that any new resource must answer. \citet{wiegand2019detection} showed that datasets built by keyword sampling produce classifiers whose apparent accuracy partly reflects the sampling procedure rather than the phenomenon. \citet{sap2019risk} showed that annotation of toxicity can encode dialectal bias, penalising varieties whose speakers are already disadvantaged. \citet{vidgen2020directions} surveyed training data across the field and found under-documentation to be the norm. \citet{rottger2021hatecheck} argued that aggregate scores on a held-out split systematically overstate model competence, and proposed functional testing as a complement. \citet{bender2018data} and \citet{gebru2021datasheets} formalised the documentation obligations that follow. We treat these as constraints on what a corpus paper may claim.

Table \ref{tab:corpora} summarises the Brazilian Portuguese resources with relevance to misogyny, whose trajectory falls into three phases. The foundation phase produced small, exploratory corpora: OffComBR \citep{pelle2017offensive} with 1{,}250 comments from the G1 news portal, framed as binary offensiveness with auxiliary type labels and baselines of $F$-measure $0.80$ for SVM and $0.75$ for Naive Bayes; and the dataset of \citet{fortuna2019hierarchically} with 5{,}668 tweets in European and Brazilian Portuguese under a binary label plus a hierarchical multilabel schema, with an LSTM baseline at $F_1 = 0.78$.

The expansion phase increased volume and platform diversity. ToLD-Br \citep{leite2020toxic} contributed roughly 21{,}000 tweets across seven categories, releasing both individual annotations and consensus levels, with a binary toxic versus non-toxic baseline of macro-$F_1 = 0.76$. HateBR \citep{vargas2022hatebr} contributed 7{,}000 expert-annotated Instagram comments from the profiles of Brazilian politicians across nine hate groups, reporting $F_1 = 0.85$ for offensive language and $F_1 = 0.78$ for hate speech. OLID-BR \citep{trajano2023olidbr} imported the OLID schema \citep{zampieri2019olid} into Portuguese, covering offensiveness, type, target and toxic spans over 6{,}354 comments later expanded to 13{,}538, with $F_1 = 0.85$ on offensiveness and $F_1 = 0.72$ on toxicity type. TuPy-E \citep{oliveira2023tupye} compiled \citet{fortuna2019hierarchically}, \citet{leite2020toxic} and \citet{vargas2022hatebr} and added new annotations, reaching 43{,}668 instances and a binary $F_1 = 0.90$.

The specialisation phase is recent and thin. MINA-BR \citep{plath2022minabr} was built from X and YouTube expressly to identify hate speech against women, and reported an SVM baseline of $F_1 = 0.57$. It is the closest antecedent to the present work. Its data were published in the research data repository of Unicamp in March 2025 \citep{plath2025minabrdata}: 6{,}002 collected comments, of which 2{,}091 were labelled by three annotators each for hate and for offensiveness, together with annotator profiles and self-rated confidence. The files are publicly downloadable, although the record states no licence and asks users to contact the depositor before use. A first independent study has since used them \citep{vicente2025detecting}. \citet{rangel2026misoginia} propose the LexMis-BR lexicon and use it to guide the collection and annotation of social media messages, releasing the artefacts that platform rules and ethical constraints permit. \corpus{} differs from both in its label definition, which separates misogyny from non-gendered toxicity, and in what accompanies the data: the triage prompt, the annotation guide, per-instance provenance, fixed partitions and a versioned release.

\begin{table}[t]
\centering
\small
\caption{Brazilian Portuguese corpora relevant to misogyny detection. ``Misogyny'' indicates whether the schema addresses the phenomenon directly, indirectly through a broader \emph{sexism} category, or not at all. $^{\dagger}$Collected comments, of which 2{,}091 are labelled.}
\label{tab:corpora}
\resizebox{\textwidth}{!}{%
\begin{tabular}{lllrll}
\toprule
\textbf{Corpus} & \textbf{Year} & \textbf{Source} & \textbf{Size} & \textbf{Misogyny} & \textbf{Availability} \\
\midrule
OffComBR & 2017 & G1 (news) & 1{,}250 & No & Public \\
\citet{fortuna2019hierarchically} & 2019 & Twitter & 5{,}668 & Indirect (sexism) & Partial \\
ToLD-Br & 2020 & Twitter & $\sim$21{,}000 & Label; 29 w/ consensus & Public \\
HateBR & 2022 & Instagram & 7{,}000 & Indirect (sexism) & Public \\
MINA-BR & 2022 & Twitter/YouTube & 6{,}002$^{\dagger}$ & Exclusive (hate) & Public since 2025 \\
OLID-BR & 2023 & Twitter/YouTube & 6{,}354--13{,}538 & Indirect (sexism) & Public \\
TuPy-E & 2023 & Compilation & 43{,}668 & Label, generalist scope & Public \\
\midrule
\textbf{\corpus{}} (this work) & 2025 & X + OLID-BR & 896 & \textbf{Exclusive} & \textbf{Public} \\
\bottomrule
\end{tabular}}
\end{table}

The gap is therefore specific rather than general. Brazilian Portuguese does not lack hate speech data; it has few open resources in which misogyny is the object of annotation rather than a facet of it, and those that exist are recent. \corpus{} does not close the volume gap. It narrows the availability gap for the specialised case, and it does so with the documentation that would allow a larger successor to be built on compatible foundations.

\section{The \corpus{} Corpus}
\label{sec:construction}

This section describes how \corpus{} was built and what the result looks like. The construction pipeline has four stages, targeted collection, automatic triage, manual annotation, and consolidation with release, and Section \ref{sec:characterization} then characterises the corpus that emerges from them. Table \ref{tab:funnel} reports the item counts at each stage, which we present as a funnel because every reduction is a decision with consequences for what the corpus can be used to measure.

\begin{table}[t]
\centering
\small
\caption{Corpus assembly funnel. All figures are as recorded during construction; the released \texttt{source} field permits exact recovery of the provenance of every instance.}
\label{tab:funnel}
\begin{tabular}{lr}
\toprule
\textbf{Stage} & \textbf{Items} \\
\midrule
Collected from X by lexicon- and target-guided scraping & 614 \\
Retained after LLM triage (predicted \emph{Toxic} or \emph{Misogynistic}) & 355 \\
Discarded by triage (predicted \emph{Neutral}) & 259 \\
Imported from OLID-BR (\texttt{sexism} and \texttt{not offensive}) & 461 \\
\midrule
Entering manual annotation and agreement analysis & 816 \\
Retained after removal of duplicates and empty strings & 815 \\
\quad of which misogynistic & 448 \\
\quad of which non-misogynistic & 367 \\
Additional OLID-BR non-offensive items added for class balance & 81 \\
\midrule
\textbf{Final corpus} & \textbf{896} \\
\quad Training partition (50/50) & 716 \\
\quad Test partition (50/50) & 180 \\
\bottomrule
\end{tabular}
\end{table}

\subsection{Collection and Triage}

X was selected as the primary source for its microblogging format, which favours condensed and highly opinionated discourse and has repeatedly proven productive for studies of gendered abuse \citep{waseem2016hateful,zampieri2019olid}. Since March 2023 the platform has charged for basic API access, which placed programmatic collection outside the budget of the project; collection was therefore performed by scraping, implemented in Python with the Twikit library. URLs and line breaks were removed at collection time, duplicates were eliminated, and randomised pauses of one to three seconds were inserted between requests after the collecting account encountered temporary blocks.

Sampling followed the target-guided strategy of \citet{waseem2016hateful}, combining observation of high-visibility profiles, journalists, artists and digital influencers, with lexicons of terms associated with misogyny and sexism drawn from MonitorA \citep{monitora2024}. The lexicon was organised along three complementary axes. The first comprises direct offensive and discriminatory terms recurrent in attacks on women, a vocabulary attested in MINA-BR \citep{plath2022minabr} and in systematic reviews of hate speech resources \citep{poletto2021resources}. The second comprises identity and social group markers relating to gender, race, regional origin and political alignment, following the observation in HateBR \citep{vargas2022hatebr} that such categories capture the intersection of misogyny with other axes of hostility. The third comprises social and behavioural context terms tied to gender roles and sexist stereotypes, a dimension emphasised by \citet{mantini2021reconhecimento} in showing that violent narratives can be constructed without any explicit slur. This procedure yielded 614 comments. To broaden coverage, 461 additional items were imported from the training split of OLID-BR \citep{trajano2023olidbr}, drawing on the \texttt{sexism} and \texttt{not offensive} categories, the former to reinforce the positive class and the latter the negative. Imported items were re-annotated under the \corpus{} guide rather than inherited, so that every label in the corpus reflects a decision made under a misogyny-specific definition; the original OLID-BR label is preserved in the \texttt{label\_source} field for auditability.

We state plainly what this sampling strategy does and does not deliver. It does not approximate the natural distribution of discourse on the platform, and no statistic computed on \corpus{} should be read as an estimate of the prevalence of misogyny on X. What it delivers is a sufficiently varied set of positive instances to support annotation and model development, at the known cost identified by \citet{wiegand2019detection}: classifiers trained on keyword-sampled data partly learn the sampling procedure.

Manual annotation is the binding constraint on corpus size, and the density of relevant material in target-guided scraping is still low enough that annotators spend most of their effort on items that will be discarded. We therefore inserted an automatic triage stage using \texttt{gemini-2.5-flash}, following the practice of documenting thresholds and prioritisation decisions explicitly \citep{trajano2023olidbr,poletto2021resources}. The model classified each of the 614 collected comments as \emph{Neutral}, \emph{Toxic} or \emph{Misogynistic}, and all items in the latter two categories, 355 in total, were forwarded to human annotation. Decoding used a temperature of $0.08$, \texttt{top\_p} of $0.8$, \texttt{top\_k} of $40$ and enforced JSON output.

The prompt was structured as a staged decision procedure rather than a category list, and this design deserves comment because it is the part of the pipeline most likely to be reused. A preliminary step asked the model to establish the overall sentiment and to check for resignified usage, the case in which a term with strong offensive potential functions as an intense vocative inside a community, as in exchanges between fans and artists, and to route such cases to \emph{Neutral} regardless of surface lexis. Only if the intent was hostile did the procedure ask whether the text was intrinsically offensive, then whether the target was a woman or the feminine in general, and finally whether the offence was motivated by gender stereotype rather than being a personal attack that happened to land on a woman. This last distinction is the one that separates \emph{Misogynistic} from \emph{Toxic} and is the conceptual centre of the whole resource. The complete prompt is released with the corpus.

The triage is efficient and it is also a source of bias. A model of this kind may under-represent subtle, ironic or coded misogyny relative to explicit forms, and may inherit associations between dialectal features and toxicity of the sort documented by \citet{sap2019risk}. Two mitigations were applied and a third is owed. Human annotation under a written guide governs every retained label, so triage cannot create a positive instance on its own. Borderline cases were reviewed. What has not been done, and what we recommend to anyone extending the corpus, is an audit of a random sample of the 259 discarded items to estimate the false negative rate of the triage stage. Without it, the recall of the pipeline with respect to the collected pool is unknown, and we do not claim otherwise.

\subsection{Annotation and Agreement}

Annotation was governed by a written guide developed for this corpus, containing operational criteria, worked examples, counter-examples, screenshots of the tool and a decision flowchart mirroring the three-step procedure used in triage. The guide is published alongside the corpus. Three non-specialist annotators participated: the first author, aged 44, university graduate; a male volunteer, aged 31, university graduate; and a female volunteer, aged 31, a fourth-semester undergraduate in systems development. Before annotation began, all three were trained on the tool and given an orientation session covering the concepts of misogyny, sexism and machismo, with time for questions. We report annotator composition in this detail because, as \citet{zeinert2021annotating} demonstrate, the identity and training of annotators is part of the definition of a misogyny corpus rather than a footnote to it, and because a panel of three, only one of whom is a woman, is a real limitation rather than a formality.

The label set had three values. \emph{Neutral} covers comments without offensive or prejudicial charge. \emph{Toxic} covers offensive or aggressive comments not specifically directed at women. \emph{Misogynistic} covers hostile comments, explicit or implicit, motivated by the feminine gender. The released corpus collapses this into the binary task by mapping \emph{Misogynistic} to $1$ and both other categories to $0$; the three-way distinction survives in the guide and is recoverable for future multiclass work. Annotation was performed in Label Studio.

Reliability was assessed with Fleiss' $\kappa$ \citep{fleiss1971measuring}, appropriate for multiple annotators over nominal categories. On the decisive question, whether a text is misogynistic, over $N = 816$ items with two categories and three raters, the coefficient reached $\kappa = 0.7880$, with observed agreement of $0.8946$ and chance agreement of $0.5029$. On the scale of \citet{landis1977measurement} this falls in the substantial band ($0.61 \leq \kappa \leq 0.80$), close to its upper boundary. Unanimity was reached on 687 items ($84.2\%$), with disagreement on 129 ($15.8\%$).

Three qualifications are in order, and we make them because a $\kappa$ in the substantial band is often reported as though it settled the matter. First, $\kappa$ is sensitive to the marginal distribution of categories, and a corpus deliberately enriched in positives has a chance agreement of $0.5029$ rather than the much higher value a naturalistic sample would produce; the coefficient is therefore not comparable to $\kappa$ values reported on differently sampled corpora \citep{artstein2008inter}. Second, an aggregate coefficient over three raters conceals pairwise structure. Third, and most informative, the disagreements are not noise. They cluster on exactly the constructions that motivate the resource: reported speech and quotation of misogynistic content, evaluative adjectives applied to a woman where gender may or may not be the ground of the evaluation, and utterances in which a gendered stereotype is invoked ironically. An item such as \emph{``Que mulher besta!!''} split the annotators two to one, and a reasonable person can defend either reading. These cases are the phenomenon, not a failure to capture it, and their retention with a \texttt{confidence\_label} field is a design choice rather than a compromise.

Disagreements were resolved by majority rule: with three annotations, the label endorsed by at least two prevailed. We follow \citet{trajano2023olidbr} in preferring this to a unanimity requirement, which would bias the positive class toward explicit instances and train models that recognise only the misogyny that announces itself. The corpus records for every instance whether its label was \texttt{unanimous} or \texttt{majority}, which allows any user to reconstruct the strict subset, to weight instances by confidence, or to evaluate models separately on the contested cases. We regard this field as one of the more useful properties of the release and encourage its use.

\subsection{Consolidation, Balance and Release}

Preprocessing lowercased the text, replaced URLs, mentions and hashtags with the tokens \texttt{[URL]}, \texttt{[USER]} and \texttt{[HASHTAG]}, normalised laughter variants (\emph{kkkk}, \emph{rsrsrs}, \emph{hahaha}) to \texttt{[LAUGHTER]}, removed duplicates and empty strings, and anonymised mentions, e-mail addresses and telephone numbers. Anonymisation was verified after partitioning by sampling with type-specific regular expressions and correcting the residual false positives and negatives by hand.

After annotation the class distribution stood at 448 misogynistic and 367 non-misogynistic, a $55$/$45$ split. Balance was achieved by drawing 81 further \texttt{not offensive} items from OLID-BR, yielding 448 per class. We flag the terminology because it matters: the additional negatives are new items drawn from an external corpus, not duplicated instances, so no instance appears twice and no leakage across the train/test boundary arises from this step. The final corpus was partitioned by stratified sampling at $80$/$20$ into 716 training and 180 test instances, each partition balanced at $50$/$50$. The partitions are fixed and released, so every number in Section \ref{sec:results} is reproducible against the same split.

Table \ref{tab:schema} gives the data dictionary. The corpus is distributed as UTF-8 CSV through the Hugging Face Hub under version tag \texttt{v1.0}, accompanied by a dataset card containing scope, field documentation, citation instructions, licence and ethical notes, in line with the documentation practices argued for by \citet{bender2018data} and \citet{gebru2021datasheets}. Users should pin the revision when loading the corpus, as we did for the experiments reported here.

\begin{table}[t]
\centering
\small
\caption{Data dictionary of the \corpus{} corpus.}
\label{tab:schema}
\begin{tabular}{p{3.1cm}p{1.0cm}p{3.5cm}p{5.2cm}}
\toprule
\textbf{Field} & \textbf{Type} & \textbf{Values} & \textbf{Description} \\
\midrule
\texttt{id} & string & hash/UUID & Stable identifier of each instance. \\
\texttt{text} & string & free text & Normalised and anonymised text. \\
\texttt{label\_misoginia} & int & $1$ misogynistic; $0$ otherwise & Final label after annotator consolidation. \\
\texttt{confidence\_label} & string & \texttt{unanimous}, \texttt{majority} & Whether the label was unanimous or decided by majority. \\
\texttt{source} & string & \texttt{twitter}, \texttt{olid-br} & Provenance: scraped from X or imported from OLID-BR. \\
\texttt{label\_source} & string & \texttt{not offensive}, \texttt{sexism}, empty & Original OLID-BR annotation; empty when \texttt{source} is \texttt{twitter}. \\
\bottomrule
\end{tabular}
\end{table}

\subsection{Characterization}
\label{sec:characterization}

We characterise the corpus with attention to the misogynistic class, since that is what the resource exists to capture. Table \ref{tab:length} reports its length statistics. The distribution is strongly right-skewed, with a mean of $21.41$ tokens against a median of $16$ and a maximum of $103$, reflecting a bimodal discursive reality: short messages consisting of a direct insult, and longer comments that build an argument for the target's illegitimacy. The two poles are not merely stylistic variants. The short pole is where lexical models perform well and where the phenomenon is least interesting; the long pole is where the corpus earns its purpose and where lexical models fail.

\begin{table}[t]
\centering
\small
\caption{Length statistics of the misogynistic class.}
\label{tab:length}
\begin{tabular}{lrrrrrrr}
\toprule
\textbf{Metric} & \textbf{Mean} & \textbf{Median} & \textbf{SD} & \textbf{Q1} & \textbf{Q3} & \textbf{Min} & \textbf{Max} \\
\midrule
Characters & 120.51 & 91.0 & 92.72 & 52.0 & 164.25 & 4 & 503 \\
Tokens & 21.41 & 16.0 & 16.83 & 9.0 & 30.0 & 1 & 103 \\
\bottomrule
\end{tabular}
\end{table}

Lexical diversity, measured as the type-token ratio, is $0.298$ over a vocabulary of $2{,}861$ types and $9{,}592$ tokens. This is consistent with the recurrent observation that hate speech corpora exhibit constrained and repetitive vocabularies, driven by the reuse of a stable stock of insults and stereotyped constructions \citep{fortuna2018survey,vargas2022hatebr,plath2022minabr,poletto2021resources}. We report the figure with an explicit caveat that is frequently omitted: TTR is a decreasing function of text length and is therefore not comparable across corpora of different sizes. The value above characterises \corpus{} and licenses no comparison with the diversity reported for larger resources. Studies wishing to compare lexical diversity across corpora should use a length-invariant measure such as MTLD \citep{mccarthy2010mtld}.

Unigram analysis of the misogynistic class is dominated by the anonymisation marker \texttt{USER}, followed by \emph{mulher}, \emph{puta}, \emph{vagabunda} and \emph{piranha}, a profile in which feminine identity is reduced to stigmas of sexuality and morality. Bigrams show intensification through lexical combination, with \emph{filha da puta} prominent, and a non-trivial role for emoji pairs, which function to intensify or to ridicule rather than to soften. The high frequency of \texttt{USER} deserves a modelling note: it is an artefact of anonymisation, it correlates with the positive class because attacks are addressed, and any classifier trained on this corpus will exploit it. Removing the token or replacing it with a constant across classes is advisable for any study interested in linguistic rather than structural signal.

Qualitative inspection confirms that misogyny in the corpus exceeds direct insult and organises itself into recognisable discursive strategies. Four recur. Sexualisation and harassment reduce the target to a sexual object, often through animalising vocatives. Intellectual disqualification denies competence on gendered grounds, as in the assertion that pregnancy makes women stupid, or in the pathologising vocabulary of \emph{louca}, \emph{doida} and \emph{desequilibrada} applied to a woman raising her voice. Stereotype enforcement polices conduct against a gendered norm, frequently through the diminutive (\emph{mulherzinha}) and through accusations of hypocrisy directed at feminism. Symbolic violence and threat range from calls for physical punishment to explicit endorsement of lethal violence, and are notable for often carrying their own meta-commentary, anticipating and pre-emptively dismissing the charge of misogyny.

The last of these is worth dwelling on, because it exposes what a lexicon cannot reach. An utterance that concludes by dismissing anyone who would call the speaker misogynistic is performing an awareness of the category while committing the act, and no keyword list will separate it from an ironic quotation of the same words. It is instances of this kind that justify a specialised corpus, and it is on instances of this kind that our own annotators disagreed most.

\section{Experimental Setup}
\label{sec:setup}

The task is binary classification of misogynistic against non-misogynistic text over the fixed partitions described in Section \ref{sec:construction}. Models were fitted on the training partition only; the test partition was held out from all hyperparameter selection and modelling decisions and used exactly once.

Four classical baselines were evaluated. Multinomial and Bernoulli Naive Bayes were fitted on word unigram and bigram counts, with the smoothing parameter explored by cross-validation on the training partition. A linear SVM (LinearSVC) and a random forest were fitted on TF-IDF representations of word unigrams and bigrams, with the regularisation parameter and the number and depth of trees respectively tuned by stratified cross-validation on the training partition. These models are included not because they are competitive but because they are the reference points against which the Brazilian literature has reported for a decade \citep{pelle2017offensive,fortuna2018survey,trajano2023olidbr}, and because their failure modes are diagnostic.

Two transformer baselines were evaluated: BERTimbau Base and BERTimbau Large \citep{souza2020bertimbau}, fine-tuned with cased tokenisation, a maximum sequence length between 128 and 256 tokens and AdamW optimisation. The Base variant was trained for up to 20 epochs with a batch size of 16, the Large variant for up to 10 epochs with a batch size of 8 owing to memory requirements. Both used early stopping with patience of 3 on validation macro-$F_1$. Base stopped at epoch 16 with its best validation score at epoch 13; Large stopped at epoch 8 with its best at epoch 5. Fine-tuning ran on Google Colab with a Tesla T4 GPU (15.8 GB), Python 3.12.12 and Transformers 4.56.1. Random seeds were fixed and library versions recorded.

The primary metric is macro-$F_1$, chosen for comparability with prior work under both balanced and unbalanced conditions \citep{fortuna2018survey,oliveira2023tupye}. We additionally report accuracy, per-class precision and recall, ROC-AUC and AUC-PR for the positive class. Because the test partition contains 180 items, a single item corresponds to $0.56$ percentage points of accuracy, and we therefore accompany every headline figure with a Wilson 95\% confidence interval \citep{wilson1927probable} and decline to interpret differences that fall inside it \citep{dror2018hitchhiker}.

Three hypotheses were stated before the experiments. H1: transformer models outperform classical baselines on macro-$F_1$ and AUC-PR. H2: among classical models, the linear SVM with TF-IDF outperforms Naive Bayes and random forest. H3: the difference between BERTimbau Base and Large is small relative to the additional computational cost.

\section{Results and Discussion}
\label{sec:results}

Table \ref{tab:classical} reports the classical baselines. The random forest with TF-IDF attains the highest macro-$F_1$ ($0.676$) and accuracy ($0.678$), the linear SVM the highest ROC-AUC ($0.747$), and Multinomial Naive Bayes the highest precision ($0.760$) at the lowest recall ($0.500$). We do not rank these models. The full range of accuracies spans $0.656$ to $0.678$, which is four test items, and the confidence intervals overlap almost completely. H2 is therefore neither confirmed nor refuted: the experiment as designed cannot distinguish a linear SVM from a random forest from a Naive Bayes classifier on this corpus, and reporting the random forest as ``best'' would be an artefact of reading a point estimate as though it were a measurement. What can be said is that TF-IDF representations appear at least as good as bag-of-words counts, that all four models cluster around $0.66$ to $0.68$ macro-$F_1$, and that a larger test partition would be required to say more.

\begin{table}[t]
\centering
\small
\caption{Classical baselines on the test partition ($n = 180$). Precision and recall refer to the misogynistic class. The bracketed interval is the Wilson 95\% CI for accuracy.}
\label{tab:classical}
\begin{tabular}{lccccc}
\toprule
\textbf{Model} & \textbf{Accuracy} & \textbf{Macro-$F_1$} & \textbf{Precision} & \textbf{Recall} & \textbf{ROC-AUC} \\
\midrule
MultinomialNB (BoW) & 0.672 \small{[0.60, 0.74]} & 0.662 & 0.760 & 0.500 & 0.717 \\
BernoulliNB (BoW) & 0.656 \small{[0.58, 0.72]} & 0.651 & 0.700 & 0.540 & 0.695 \\
LinearSVC (TF-IDF) & 0.672 \small{[0.60, 0.74]} & 0.671 & 0.700 & 0.600 & 0.747 \\
Random Forest (TF-IDF) & 0.678 \small{[0.61, 0.74]} & 0.676 & 0.710 & 0.600 & 0.731 \\
\bottomrule
\end{tabular}
\end{table}

What the table does support is a substantive observation about the shape of the errors. Every classical model is precision-oriented and recall-poor, with Multinomial Naive Bayes reaching only $0.500$ recall while holding $0.760$ precision. These models find the misogyny that declares itself lexically and miss the rest. Given that a substantial share of the positive class consists of the implicit, ironic and stereotype-driven constructions catalogued in Section \ref{sec:characterization}, a recall between one half and $0.60$ is consistent with the ceiling of a purely lexical decision rule, which is the limitation that motivated the resource.

Table \ref{tab:transformer} reports the transformer baselines. Both variants reach a test macro-$F_1$ of $0.833$ and, notably, produce identical confusion matrices: 76 true negatives, 14 false positives, 16 false negatives and 74 true positives, for an accuracy of $150/180 = 0.833$ with a Wilson 95\% CI of $[0.772, 0.881]$. Precision on the misogynistic class is $0.841$ and recall $0.822$. The gap over the best classical baseline is $0.157$ macro-$F_1$, and the accuracy confidence intervals do not overlap ($[0.772, 0.881]$ against $[0.610, 0.742]$), so H1 is supported. The gain is larger in recall than in precision: recall on the misogynistic class rises from $0.600$ to $0.822$ while precision rises from $0.710$ to $0.841$. This pattern is consistent with the transformers recovering implicit positives that the lexical models cannot represent, without paying for it in false alarms, although a per-item error analysis would be needed to confirm it.

\begin{table}[t]
\centering
\small
\caption{BERTimbau baselines: performance and cost. Epoch denotes the best validation epoch selected by early stopping.}
\label{tab:transformer}
\begin{tabular}{lccccc}
\toprule
\textbf{Model} & \textbf{Epoch} & \textbf{Macro-$F_1$} & \textbf{ROC-AUC} & \textbf{AUC-PR} & \textbf{Train time (min)} \\
\midrule
BERTimbau Base & 13 & 0.8333 & 0.8993 & 0.9107 & \textbf{8.49} \\
BERTimbau Large & 5 & 0.8333 & \textbf{0.9101} & \textbf{0.9193} & 22.22 \\
\bottomrule
\end{tabular}
\end{table}

The comparison between Base and Large is more interesting for what it forecloses. The two models have the same macro-$F_1$ and the same aggregate confusion matrix, so no paired test can separate them: because both make the same number of errors on the same 180 items, the items that only Base misclassifies and the items that only Large misclassifies are necessarily equal in number, and the McNemar statistic \citep{mcnemar1947note} is zero whatever the per-item predictions. The Large variant holds a marginal advantage in the threshold-free metrics, $+0.011$ ROC-AUC and $+0.009$ AUC-PR, which suggests slightly better calibrated ranking without any effect on the decisions actually taken at the operating threshold. Against this, it costs $22.22$ minutes of training against $8.49$, a factor of $2.6$, with a correspondingly slower inference. H3 is supported, and the practical recommendation is unambiguous: on a corpus of this size the Base variant is the rational choice, and the capacity of the Large variant is idle because the training signal is too small to use it.

Three qualifications attach to the figure of $0.833$ before it can be used for anything: what it predicts about deployment, what part of it may not be measuring misogyny at all, and what it would have been under a different resolution rule. The first is prevalence. A macro-$F_1$ of $0.833$ on a balanced test set is a reasonable result for a corpus of 716 training instances and a poor predictor of behaviour in deployment, because \corpus{} is balanced at $50\%$ positives by construction while the base rate of misogynistic content in an unfiltered stream is far lower. Precision is a function of that base rate, and the function is unforgiving. Take the observed operating point of BERTimbau Base: true positive rate $74/90 = 0.822$, false positive rate $14/90 = 0.156$. Holding these fixed and varying the prevalence $\pi$, precision is
\[
\text{Precision}(\pi) = \frac{\pi \cdot 0.822}{\pi \cdot 0.822 + (1 - \pi) \cdot 0.156}.
\]
At $\pi = 0.5$ this returns the observed $0.84$. At $\pi = 0.05$, a generous estimate for a stream of replies to a public figure, it returns $0.22$. At $\pi = 0.01$ it returns $0.05$: nineteen of every twenty flags would be false. The model has not become worse; the metric was never measuring what a deployment cares about. This is not a defect peculiar to \corpus{}, and it is not an argument against balancing, which is defensible for a corpus of this size where an unbalanced split would leave too few positives to learn from. It is an argument for stating the consequence explicitly rather than allowing a balanced macro-$F_1$ to circulate as a deployment claim. Anyone building a moderation system from this corpus must calibrate the threshold against the prevalence of their own stream, must expect the false positive rate to dominate, and should treat a $0.156$ FPR as the number to reduce.

The second qualification concerns provenance. Roughly $60\%$ of the corpus derives from OLID-BR and roughly $40\%$ from scraping, and this split is not orthogonal to the label. The scraped portion was collected using misogyny lexicons and therefore contributes disproportionately to the positive class; the OLID-BR portion contributes the \texttt{not offensive} items, including all 81 of the balancing top-up, and therefore disproportionately to the negative class. A classifier can reduce its loss by learning to discriminate scraped-tweet register from OLID-BR register, and to the extent that it does so, part of the $0.833$ is measuring provenance rather than misogyny. We do not know how large that part is, and we say so. We also make it measurable: the \texttt{source} field is released precisely so that this can be quantified, and we recommend three checks to anyone using the corpus for modelling claims. Train on the corpus and evaluate on the scraped subset alone and the OLID-BR subset alone, and compare. Train a classifier to predict \texttt{source} from \texttt{text} and inspect its accuracy; anything well above chance quantifies the confound directly. Evaluate on the \texttt{unanimous} and \texttt{majority} subsets separately, since the confound and the difficulty are likely correlated. Version 2 of the corpus should be sourced so that provenance and label are independent, which chiefly means annotating enough scraped negatives to make the OLID-BR top-up unnecessary.

The third qualification runs in the opposite direction. Where provenance may inflate the headline figure, the resolution rule deflates it. The $129$ disagreements, $15.8\%$ of annotated items, are in our view the most informative part of the resource, and they are not distributed randomly: they concentrate on reported speech, on evaluative predicates whose gendered grounding is arguable, and on irony. These are the constructions on which any deployed system will be contested, and on which the classical baselines fail almost completely. Retaining them under majority rule with a recorded \texttt{confidence\_label}, rather than discarding them or requiring unanimity, follows \citet{trajano2023olidbr} and is consequential. Requiring unanimity would have produced a corpus of $687$ items whose positive class was systematically more explicit, on which every model would have scored higher, and from which every model would have learned less. The reported $0.833$ is therefore lower than it would have been under a stricter rule.

One comparison remains, and we do not draw it. The reported baseline for MINA-BR is $F_1 = 0.57$ for SVM \citep{plath2022minabr}, against $0.671$ for our linear SVM and $0.833$ for BERTimbau Base, and it would be straightforward, and wrong, to present this as evidence that \corpus{} supports better models. The two corpora differ in class balance, in sampling strategy, in platform composition, in annotation guideline, in annotator panel, in label definition and in test partition. A difference in $F_1$ across corpora measures the joint effect of all of these and isolates none. More concretely, \corpus{} is balanced by construction and MINA-BR is not; $F_1$ on the positive class is a function of prevalence, as the calculation above makes plain, so the two figures are not on a common scale. The experiment that would license a comparison, evaluating the same model on both corpora or transferring a model trained on one to the other, has become feasible now that the MINA-BR data are public \citep{plath2025minabrdata}; we have not run it, and it is the most immediate follow-up to this work. We therefore report the MINA-BR figure for context and make no claim of superiority.

For practitioners, three findings transfer. The transformer gap is large, roughly 16 macro-$F_1$ points, with the larger share of the gain in recall on the misogynistic class; none of the lexical baselines exceeded $0.60$ recall on this corpus. Model capacity beyond BERTimbau Base is not usable at this data scale, since the Large variant costs $2.6\times$ the training time for a difference indistinguishable from zero. And the \texttt{[USER]} token is a structural artefact correlated with the positive class that should be neutralised in any study interested in linguistic signal.

\section{Limitations and Threats to Validity}
\label{sec:limitations}

We group the limitations by the validity they threaten.

Construct validity is threatened primarily by the triage stage, which may systematically favour explicit hostility and under-represent subtle, ironic or contextual misogyny. The 259 discarded items were not audited, so the false negative rate of triage is unmeasured, and the corpus therefore under-represents whatever \texttt{gemini-2.5-flash} does not recognise. A related threat comes from the annotator panel: three annotators, one of them a woman, is a minimal rather than an adequate basis for defining what counts as misogyny, and \citet{zeinert2021annotating} give reason to expect that a differently composed panel would draw the boundary differently. Finally, collapsing the three-way \emph{Neutral}/\emph{Toxic}/\emph{Misogynistic} distinction into a binary label discards the information that separates gendered from non-gendered hostility, which is arguably the most interesting distinction the guide encodes.

Internal validity is threatened by the provenance confound described in Section \ref{sec:results}, by the choice of triage prompt, model and thresholds, which shaped the pool sent to annotation and whose sensitivity was not explored, and by the fact that the balancing top-up drew all 81 additional negatives from a single external source. Hyperparameter selection was performed on the training partition and the test partition was used once, which protects against the most common form of optimism, but the test partition is small enough that its idiosyncrasies are consequential.

External validity is threatened by the single platform, single language variety and single collection window, by the lexicon- and target-guided sampling that departs by design from natural prevalence \citep{wiegand2019detection}, and by the artificial class balance whose effect on precision we quantified above. \corpus{} supports research on the linguistic characterisation of Brazilian Portuguese misogyny and on comparative modelling under a fixed protocol. It does not support estimates of prevalence, claims about platforms other than X, or claims about model performance in deployment.

Statistical conclusion validity is bounded by the test partition of 180 items. Differences below roughly 7 accuracy points are not distinguishable at this sample size, which is why we ranked no classical model and separated no BERTimbau variant. Any future work reporting improvements on \corpus{} should report confidence intervals or paired significance tests and should treat sub-point gains as noise \citep{dror2018hitchhiker}.

Finally, and encompassing all of the above, the volume of 896 instances bounds what the corpus can support. It is adequate for fine-tuning a pre-trained Portuguese encoder, as the results show, and it is inadequate for training from scratch, for multiclass or multilabel extension, for meaningful subgroup analysis, or for functional testing of the kind advocated by \citet{rottger2021hatecheck}. We present \corpus{} as a documented starting point with a protocol that can be scaled, not as a benchmark of record.

\section{Conclusion}
\label{sec:conclusion}

This paper introduced \corpus{}, an openly available corpus dedicated specifically to misogyny detection in Brazilian Portuguese and released together with its full construction pipeline. It comprises 896 balanced instances collected from X by lexicon- and target-guided scraping and supplemented from OLID-BR, triaged by a documented large language model stage, and annotated by three trained annotators under a published guide, reaching a Fleiss' $\kappa$ of $0.788$ with $89.5\%$ observed agreement. Baselines span classical models clustering around $0.66$ to $0.68$ macro-$F_1$ and fine-tuned BERTimbau variants reaching $0.833$, with the Base variant matching the Large at $2.6\times$ less training cost.

The contribution is not the score. It is a resource in which every step from collection to release is inspectable, in which the label of every instance can be traced to its provenance and to the confidence with which annotators assigned it, and whose known weaknesses have been named and made measurable rather than smoothed over. Specialised resources for Brazilian Portuguese misogyny detection have only recently started to become available to researchers outside the groups that built them. \corpus{} is open from its first version, and it is accompanied by the baselines, the protocol and the partitions against which any successor can be measured, including a cross-corpus evaluation against MINA-BR.

Four directions follow. The corpus should be expanded across platforms, time windows and topical domains, with sourcing designed so that provenance is independent of label. The annotation should be enriched with the multilevel structure the guide already implies, recovering the three-way distinction, adding severity, target and discursive strategy, and adding span-level annotation to support interpretability. The triage stage should be audited by re-annotating a random sample of discarded items, which would convert the pipeline's recall from an unknown into a measured quantity. And robustness and fairness should be evaluated directly, through functional tests, calibration analysis, obfuscation resistance and cross-platform transfer, rather than inferred from an aggregate score on a balanced split.

\section*{Ethics Statement}

The corpus consists of publicly posted social media content collected without interaction with users. All texts were normalised and anonymised, with mentions, e-mail addresses, telephone numbers and URLs replaced by generic tokens, and anonymisation was verified by sampling after partitioning. No user identifiers, profile metadata or conversation threads are released. The corpus contains material that is offensive by construction and is distributed for research on the detection and mitigation of gendered hostility; the dataset card states this scope and the conditions of use.

Annotators were informed of the nature of the material before consenting to participate, received an orientation session, and worked at their own pace. We note, for those replicating this protocol, that sustained exposure to misogynistic content imposes a real burden on annotators, that this burden is not distributed evenly across a panel, and that a female annotator reading attacks on women is not doing the same work as her colleagues.

Systems trained on this corpus can cause harm if deployed as automatic moderators. The precision analysis in Section \ref{sec:results} shows that a model performing at $0.84$ precision on the balanced test set would fall to roughly $0.22$ at a $5\%$ base rate. We therefore recommend that any application of this resource operate as decision support for human moderators rather than autonomously, and that thresholds be calibrated against the prevalence of the target stream.

\section*{Data and Code Availability}

The corpus is available at \url{https://huggingface.co/datasets/fabiopassos/miso-br} under version tag \texttt{v1.0}, with a dataset card documenting scope, fields, licence, citation and ethical notes. It loads through the Hugging Face \texttt{datasets} library with the \texttt{revision} argument pinned to \texttt{v1.0}, which is the revision used for every experiment reported here. The annotation guide is available at \url{https://fasp-school.github.io/nlp-misoginia/}. The triage prompt, fine-tuned models and experimental scripts are released alongside the corpus.

\bibliography{references}

\end{document}